\documentclass[a4paper,fleqn]{cas-sc}

\usepackage[authoryear,longnamesfirst]{natbib}
\usepackage{makecell}
\usepackage{threeparttable}
\UseRawInputEncoding

\def\tsc#1{\csdef{#1}{\textsc{\lowercase{#1}}\xspace}}
\tsc{WGM}
\tsc{QE}

\begin{document}
\let\WriteBookmarks\relax
\def\floatpagepagefraction{1}
\def\textpagefraction{.001}

\shorttitle{DINF:Dynamic Instance Noise Filter for Occluded Pedestrian Detection}    

\shortauthors{Li Xiang, He Miao, Luo Haibo, Xiao Jiajie}  

\title [mode = title]{DINF:Dynamic Instance Noise Filter for Occluded Pedestrian Detection}  



%

\author[1,2,3,4]{Xiang Li}
\cormark[0]
\credit{Conceptualization, Methodology, Software, Validation, Formal analysis, Investigation, Data Curation, Writing-Original Draft, Visualization}

\author[1,2,3]{Miao He}
\credit{Writing-Review \& Editing}

\author[1,2,3]{Haibo Luo}
\cormark[1]
\credit{Supervision, Resources, Writing-Review \& Editing}

\author[5]{Jiajie Xiao}
\credit{Writing-Review \& Editing}

\cortext[1]{Corresponding author}

\affiliation[1]{organization={Key Laboratory of Opto-Electronic Information Processing,
Chinese Academy of Sciences}, 
            city={Shenyang},
            postcode={110016}, 
            state={Liaoning},
            country={China}}
\affiliation[2]{organization={Shenyang Institute of Automation, Chinese Academy of Sciences},
            city={Shenyang},
            postcode={110016}, 
            state={Liaoning},
            country={China}}
\affiliation[3]{organization={Institutes for Robotics and Intelligent Manufacturing, Chinese Academy of Sciences},
            city={Shenyang},
            postcode={110169}, 
            state={Liaoning},
            country={China}}
\affiliation[4]{organization={University of Chinese Academy of Sciences},
            city={Beijing},
            postcode={100049}, 
            country={China}}

\affiliation[5]{organization={Trip.com Group},
            city={Shanghai},
            postcode={200335}, 
            country={China}}


\begin{abstract}
Occlusion issue is the biggest challenge in pedestrian detection. RCNN-based detectors extract instance features by cropping rectangle regions of interest in the feature maps. However, the visible pixels of the occluded objects are limited, making the rectangle instance feature mixed with a lot of instance-irrelevant noise information.  Besides, by counting the number of instances with different degrees of overlap of CrowdHuman dataset, we find that the number of severely overlapping objects and slightly overlapping objects are unbalanced, which may exacerbate the challenges posed by occlusion issues.Regarding to the noise issue, from the perspective of denoising, an iterable dynamic instance noise filter (DINF) is proposed for the RCNN-based pedestrian detectors to improve the signal-noise ratio of the instance feature. Simulating the wavelet denoising process, we use the instance feature vector to generate dynamic convolutional kernels to transform the RoIs features to a domain in which the near-zero values represent the noise information. Then, soft thresholding with channel-wise adaptive thresholds is applied to convert the near-zero values to zero to filter out noise information. For the imbalance issue, we propose an IoU-Focal factor (IFF) to modulate the contributions of the well-regressed boxes and the bad-regressed boxes to the loss in the training process, paying more attention to the minority severely overlapping objects. Extensive experiments conducted on CrowdHuman and CityPersons demonstrate that our methods can help RCNN-based pedestrian detectors achieve state-of-the-art performance.
\end{abstract}



\begin{keywords}
 Occlsuion detection \sep Pedestrian detection \sep Denoising \sep Object detection \sep Convolutional nerual networks
\end{keywords}

\maketitle

\section{Introduction}

Pedestrian detection, as a fundamental component of intelligent surveillance systems, automatic driving, etc., has been significantly developed with the rise of deep learning. Current state-of-the-art (SOTA) pedestrian detectors are almost RCNN-based detectors, such as \cite{CrowdDet}, \cite{IterDet}, \cite{PBM},\cite{Occlusion-Aware}. However, occlusion is still the most challenging issue for pedestrian detection in the crowd.

\begin{figure}[h]%
\centering
\includegraphics[width=0.3\textwidth]{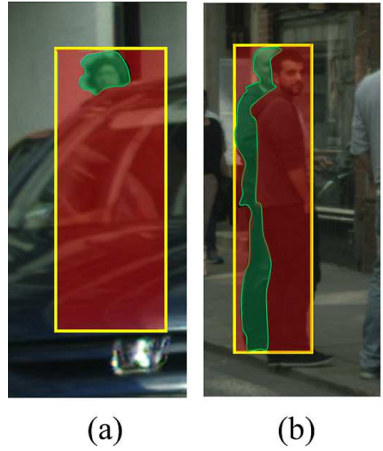}
\caption{Illustration of (a) occluded by other stuff and (b) occluded by other pedestrian}\label{fig1}
\end{figure}

Typically, there are two types of occlusion patterns: 1) occluded by other stuff; 2) occluded by other pedestrians, as shown in Fig1. For the occluded pedestrian, its visible pixels in the image are insufficient, causing little related information can be extracted. In RCNN-based detectors, the Region of Interests Pooling (RoI Pooling) is used to extract the instance features. In occlusion scenes, the RoI of the occluded pedestrian, whose most parts are occupied by other irrelevant pixels, always contains much noise information. For the first type of occlusion pattern, as shown in Fig1(a), the noise comes from the surrounding, which is generally easier to be distinguished from the pedestrian target. However, the pedestrian information might be covered up if the surrounding noise is too much. For the second occlusion pattern, as shown in Fig1(b), the feature of the occluded pedestrian fused much information of another pedestrian, which means that the noise basically comes from other pedestrian targets, i.e., they are the noise source mutually. The second occlusion type typically makes the instance feature of the occludee heavily similar to the occluder, which will generate unprecise and ambiguous prediction boxes between them and, accordingly, causes troubles to the anchor-based detectors in the non-maximum suppression (NMS). For RCNN-based detectors, the RoIs are rectangle tensors that are hard-cropped from the feature pyramid, making it inevitable to introduce noise information, not to mention that the RoI proposals usually are too rough to surround the targets precisely.\cite{Occlusion-Aware} proposed an occlusion-aware RoI pooling unit, trying to depart each proposal box into five rectangle sub-proposals, then pooling each sub-proposals and filtering out the ones containing occluder information by reweighting them. The occlusion-aware RoI pooling still operates a hard-crop mode to extract features of occludees, which can not avoid the introduction of noise. \cite{Guided-Attention} proposed an attention guidance module to help to focus on the visible parts of each occludee in RoIs. Although the attention guidance module operates in a soft mode, the extra visible-part annotations are required to maximize its effectiveness. 

In this paper, we try to think about the occlusion issue from the perspective of denoising. Inspired by the \cite{Shrinkage-Networks} and \cite{Dynamic-filter-networks} \cite{Sparse-RCNN}, we propose a dynamic instance noise filter (DINF), which operates in a soft mode to filter out noise information. Specifically, we use each instance feature to generate dynamic convolutional kernels and interact with its corresponding RoI feature. These dynamic convolutional kernels are customized according to each instance, aiming to make the interaction mechanism capable of directional noise removal. The structure of the dynamic interaction module is similar to the dynamic instance interactive head in \cite{Sparse-RCNN}. The main difference is that we replace the rectified linear unit (ReLU) with the deep learning-based soft thresholding filter, simulating the wavelet denoising to explicitly remove the noise information in the input instance feature. Our DINF can be integrated into any RCNN-based detectors. And just as the performance of all other RCNN-based detectors can be enhanced in an iterative structure, our DINF can also be integrated into an iterative noise filter network (INFN) to boost performance constantly. 

\begin{figure}[h]%
\centering
\includegraphics[width=0.5\textwidth]{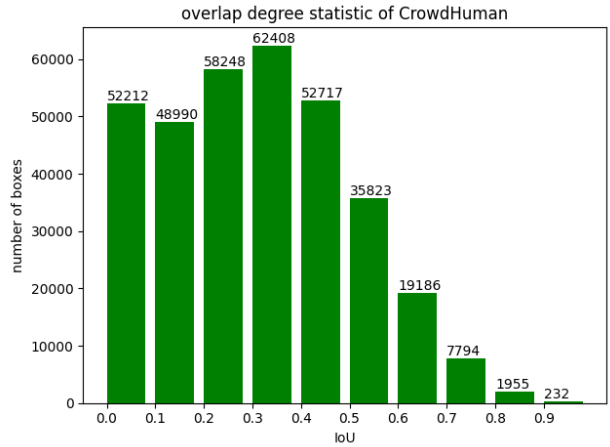}
\caption{Overlap degree statistic of CrowdHuman dataset}\label{fig1}
\end{figure}

In addition, the imbalance of training samples will significantly impact the performance of detectors. As discussed in \cite{Focal-Loss}, the balance between easily-classified samples and hard-classified samples is as important as the balance between positive and negative samples. For the second occlusion pattern, it is obvious that the more two pedestrians overlap, the more difficulty distinguishing them. However, according to our statistics on the instance overlapping degree in CrowHuman\cite{CrowdHuman} dataset, the instances of pedestrians whose Intersection over Union (IoU) with other pedestrians is in the interval [0,0.5] are the majority, as shown in Fig2. Because of the unbalance issue, the hardly-distinguished but minority samples are easily buried by others and might not be noticed in the training process. 

In this paper, following the same idea as the focal loss, we propose an IoU focal factor (IFF) for the imbalance between easily-regressed and hard-regressed samples. For each sample, according to its regression quality, i.e., the IoU value of its prediction box and ground truth box, we generate a weight to modulate the contribution of each sample to the loss. The closer the IoU tends to be 1, the smaller the corresponding IFF will be, and the more the loss of the well-regressed boxes will be reduced. Accordingly, the hard-regressed samples will contribute more to training loss in the optimization process. 

Our contributions in this paper are summarized as follows:
\begin{itemize}
\item We propose to handle the occlusion issue of pedestrian detection from the perspective of denoising and propose a dynamic instance noise filter, which can be integrated into any RCNN-based detectors.
\item We propose an iterative noise filter network (INFN) to constantly increase the signal-noise ratio (SNR) of instance features by iterating the DINF multiple times.
\item We indicate the imbalance between easily-regressed and hard-regressed samples in pedestrian detection and propose an IoU focal factor, which is simple but efficient.
\item We achieve SOTA results on CrowdHuman by integrating our INFN into other RCNN-based pedestrian detectors. Compared with the Faster-RCNN baseline, our approaches achieve 1.85\% AP, 1.15\% MR$^{-2}$, and 1.61\% JI gains. Besides, the CrowdDet equipped with INFN achieves the optimal performance among all box-based methods.

\end{itemize}

\section{Related Works}

\paragraph{Generic Object Detection} With the assistance of convolutional neural networks (CNN), the performance of object detectors has been enormously boosted compared with the traditional methods. RCNN\cite{RCNN} is the first detector to introduce the CNN into object detection and initiates the two-stage methods, which is a kind of mainstream method type nowadays. It extracts instance features by cropping rectangle regions from the deep feature maps. The Fast-RCNN \cite{Fast-RCNN} and Faster-RCNN \cite{Faster-RCNN} were proposed subsequently to overcome the low inference speed of RCNN, in which the Faster-RCNN proposed the region proposal network (RPN) to generate high-quality proposals rather than randomly generated using \cite{Selective-Search}. The Faster-RCNN lays the framework foundation for the two-stage detectors. In order to share the later computation, the cropped tensors need to be pooled to the same size. Many pooling methods were proposed, such as \cite{RoI-Warp, Precise-RoI-Pooling, RoI-Align, PS-RoI-Pooling}. To further boost the inference speed, one-stage methods are proposed, such as SSD\cite{SSD}, DSSD\cite{DSSD}, YOLO series\cite{YOLOv1,YOLOv2,YOLOv3,YOLOv4, Scaled-YOLOv4}, and RetinaNet\cite{Focal-Loss}. Rather than extracting each instance feature by cropping the feature maps, one-stage methods pave dense anchor boxes across the feature maps to capture as many objects with random locations, scales, and shapes as possible.Compared with the two-stage methods, one-stage methods face an imbalance issue between positive and negative samples, which influences the performance seriously. \cite{Focal-Loss} proposed a focal loss to handle this issue, which overcomes the imbalance of positive and negative as well as the imbalance of easy and hard classified samples. With the focal loss, the one-stage detector RetinaNet became the first one-stage detector that achieves comparable performance as the two-stage detectors. Later, some researchers tried to discard the anchors and proposed anchor-free methods such as CornerNet\cite{Cornernet},CenterNet\cite{CenterNet},FCOS\cite{FCOS} to avoid the shortcomings brought by the manually designed anchor boxes. But in terms of pedestrian detection, the two-stage methods still outperform the anchor-based one-stage methods and anchor-free methods.
\paragraph{Occlusion Handling for Pedestrian Detection} As a basic task of computer vision, pedestrian detection keeps receiving interest with a wide range of applications. Although the CNN-based methods have dominated pedestrian detection, the occlusion issue is still challenging for the current detectors. The visible part of occludee is limited, and its appearance is similar to the occluder, making it hard to be distinguished from the occluder in the feature space. Some methods \cite{Occlusion-Aware,A-Structural-Filter-Approach-to-Human,A-discriminative-deep-model-for-pedestrian-detection-with-occlusion-handling,Deep-Learning-Strong-Parts-for-Pedestrian-Detection,Joint-Deep-Learning-for-Pedestrian-Detection,Handling-Occlusions-with-Franken-Classifiers,Learning-to-Integrate-Occlusion-Specific-Detectors-for-Heavily-Occluded-Pedestrian-Detection,Multi-label-Learning-of-Part-Detectors-for-Heavily-Occluded-Pedestrian-Detection}use part-based models, which depart a pedestrian object into several parts and fuse the part detection results as the final output. Generally, each part is represented as a rectangle region, making it inevitable to introduce noise. These part-based methods can be regarded as hard-crop methods. In contrast, \cite{Guided-Attention} introduced the attention mechanism to boost the SNR in a soft way. The author found that specific parts of the human body will activate different channels of the RoI features, so they used channel-wise attention \cite{SENet} to focus on the visible part of the occludee. \cite{Attention-guided-neural-network} divided each image into several non-overlapping sub-images and used CNN to extract deep features. After that, the attention mechanism was adopted to select the features representing the body parts of a pedestrian. Then, the selected features were sent to a recurrent neural network to generate final predictions. Besides the explicit occlusion feature extraction, some other papers try to design proper loss functions. \cite{Occlusion-Aware} proposed an AggLoss to improve the compactness of the prediction boxes. \cite{Repulsion-Loss} inserted a push force between the prediction boxes belonging to different objects to make them more distinguishable. 

The occlusion will also cause trouble with the non-maximum suppression (NMS). \cite{CrowdDet} proposed a novel concept that predicts multi-instances based on the same proposal. At the same time, a set-NMS was proposed to suppress the boxes with IoU higher than the threshold but belonging to different objects with the current candidate prediction, which can preserve the heavily overlapping pedestrian not to be suppressed by mistake. \cite{Beta-RCNN} pictured a pedestrian by a beta distribution representation generated from corresponding full-body and visible boxes. Compared with the bounding box representation, distribution representations are more distinguishable during the NMS process. Similarly, \cite{PBM, Visibility-Guided-NMS} utilized the additional visible annotations of each instance to remove the full prediction boxes because the visible parts hardly overlap with others. \cite{Adaptive-NMS} focused on the NMS threshold and tried to customize the IoU threshold for different instances. \cite{NOH-NMS} attempted to perceive the maximum overlap object for each prediction box by outputting the relative information between them to avoid false suppression of true positive predictions during the NMS process.

\section{Methods}

In this section, we first introduce the details of the proposed DINF and how to use it to structure the INFN. Secondly, the proposed IFF is presented. 

\subsection{Dynamic Instance Noise Filter}

RoI pooling is the signature component of RCNN-based detectors. However, since the shapes of objects are rarely rectangular, not to mention the proposals of RoI pooling are always imprecise, making the rectangle pooling manner inevitably introduce noise. In occlusion scenes, the occludee and occluder will mutually occupy partial regions of mutual proposals so that they will become each other's noise. In this paper, we propose to handle the occlusion issue from the perspective of denoising.

\paragraph{Theoretical Background} Soft thresholding is a denoising method that was initially proposed in \cite{De-noising-by-soft-thresholding}. In general soft denoising methods, the raw signal is first transformed to a domain where the near-zero numbers represent the noise, and the large amplitude numbers represent the useful signal. Then, soft thresholding is applied to convert the near-zero numbers to zeros. Eventually, the filtered signal will be inverted back to the original domain. The soft thresholding can be expressed as follows:

\begin{equation}
y=\left\{
\begin{array}{cl}
x-\tau, &  x > \tau \\
0,  &  -\tau \le x \le \tau \\
x+\tau, &  x<-\tau \\
\end{array} \right.
\end{equation}

In the above process, two key tasks ensure a good denoising performance: 1) design a suitable filter that can transform useful information into positive or negative values with large amplitude and noise information into near-zero values. 2) design an appropriate threshold $\tau$ for removing noise information and preserving useful information as much as possible. In the past, designing such an ideal filter and threshold was challenging and required extensive knowledge of signal processing. Nowadays, depending on automatically learning process using a gradient descent algorithm, deep learning provides a new  approach to finding a suitable filter and a threshold by training the neural networks. Simulating the classical wavelet denoising method, \cite{Shrinkage-Networks}proposed using learnable convolutional kernels to imitate the ideal wavelet filter and used fully-connected(FC) networks to generate adaptive thresholds for input features. As shown in Fig3(a), the global average pooling (GAP) is first applied to the absolute values of the input feature $x$ to get a feature vector $\bar{\left |x \right |}$. Then $\bar{\left |x \right |}$ is fed into two FC layers with the same output dimension as the input and output a scaling parameter $z$. A sigmoid function is then applied to the output of the FC layers $z$ to scale the scaling parameter to the range of $(0, 1)$:

\begin{equation}
\alpha =\frac{1}{1+e^{-z}} 
\end{equation}
where $\alpha$ is the final scaling parameter, which is then multiplied with the  $\bar{\left |x \right |}$ to obtain the final threshold $\tau$ :
\begin{equation}
\tau =\bar{\left |x \right |} \odot \alpha
\end{equation}
where $\odot$ is the element-wise multiplication. Since the $\tau$ is a 1-D feature with the same dimension as the input feature $x$, the soft thresholding in each channel is variant. 
\begin{figure}[tp]%
\centering
\includegraphics[width=1.0\textwidth]{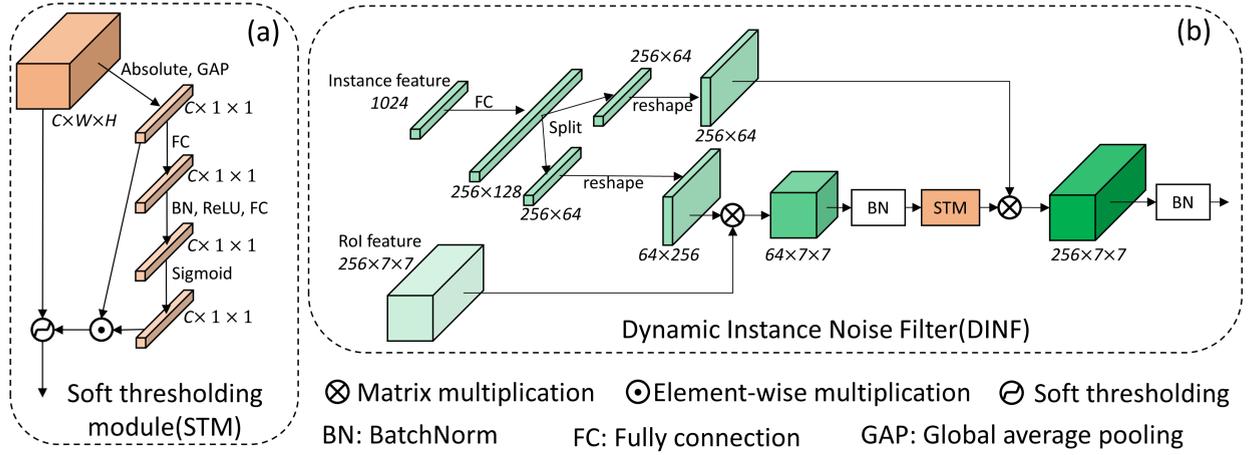}
\caption{Illustration of (a) the soft thresholding module and (b) the dynamic instance noise filter}\label{fig3}
\end{figure}

\paragraph{DINF} From the perspective of denoising, we propose the DINF to filter out the irrelevant information from the RoI features, as shown in Fig3(b). In DINF, we take the feature vector fed to the prediction head as the instance feature, which has been filtered by two FC layers from the RoI features and is of relatively higher SNR than the RoI feature. Inspired by \cite{Sparse-RCNN,Dynamic-filter-networks}, the instance feature is used to generate dynamic convolutional kernels. As Fig3(b) shows, in DINF, the instance feature is first sent to a FC layer to output a feature vector with the dimension of $256\times 128$. Then, the feature vector is split into two sub-feature vectors with a dimension of $256\times 64$. These two sub-feature vectors are reshaped as two matrixes with the size of $256\times 64$ and $ 64\times256$, respectively. The two matrixes can be seen as two dynamic 1$\times$1 convolutional kernels with output channels of 64 and 256. In DINF, the first dynamic convolutional kernel is used to transform the input RoI feature into the domain in which the near-zero values are noise, and the second one is used to invert the filtered features back to the original domain. The DINF takes instance features with different occlusion degree and different occlusion type as inputs, meaning that the input instance features have different SNR levels. The dynamic convolutional kernels customized for variant instance features ensure adaptive simulative wavelet filter for each input instance feature. The dynamic interaction structure of DINF is inspired by the dynamic instance interactive head in \cite{Sparse-RCNN}, which claimed the dynamic interaction between the RoI feature and its corresponding instance feature could filter out ineffective bins of the input RoI feature. But unlike the dynamic instance interactive head using ReLU as the activation function, we adopt the STM to conduct the soft denoising process explicitly. The comparison of soft thresholding and ReLU are illustrated in Fig4. Soft thresholding focuses on the signal with large amplitudes rather than with positive values. 
\begin{figure}[h]%
\centering
\includegraphics[width=0.8\textwidth]{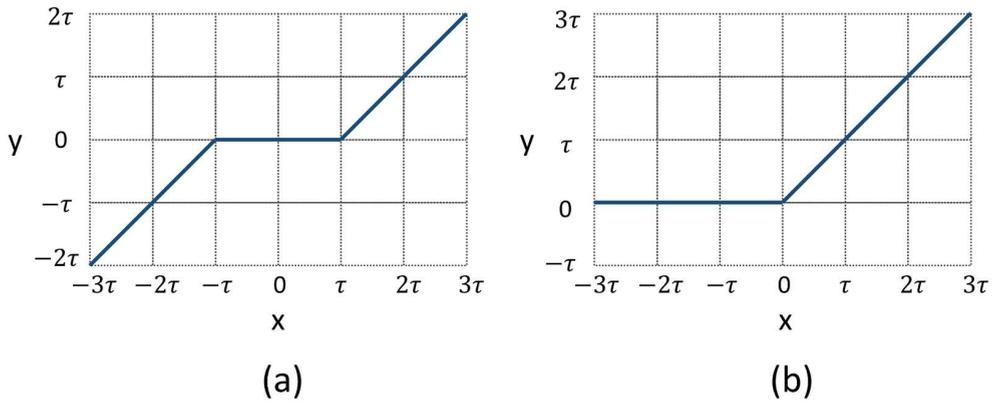}
\caption{Illustration of (a) soft thresholding and (b) the ReLU}\label{fig4}
\end{figure}

In summary, following the principle of wavelet denoising, through the dynamic convolutional kernel, our DINF first transforms the RoI feature into the domain in which the noise signal values are near-zero numbers. The soft thresholding is then applied to adaptively convert the near-zero values to zero to remove the noise explicitly.Subsequently, the denoised features are transformed back to the original domain by another dynamic convolutional kernel. 

\begin{figure}[h]%
\centering
\includegraphics[width=1.0\textwidth]{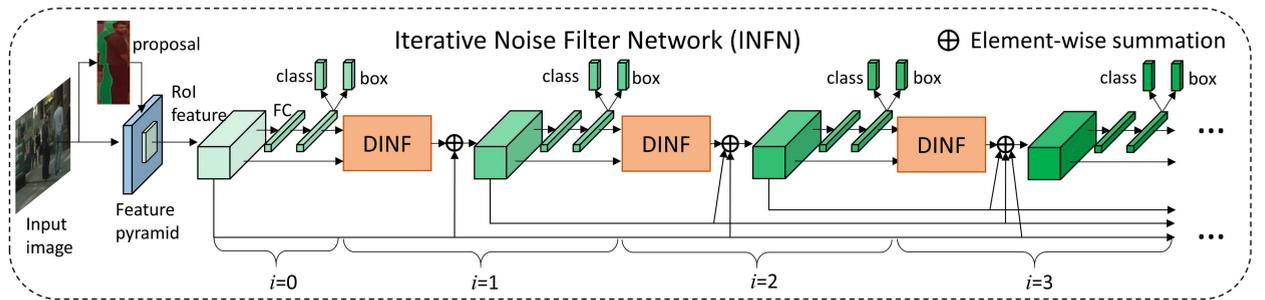}
\caption{Illustration of the Iterative noise filter network}\label{fig5}
\end{figure}

\subsection{Iterative Noise Filter Networks}
The iterative structure is a regular method to boost the performance of RCNN-based detectors. Following the same principle, we integrate the DINF into an iterative noise filter network (INFN) to further increase the SNR by iterating the DINF multiple times, as shown in Fig 5. Rather than using the prediction boxes to crop new RoI features, we operate the iteration at the feature level. A perfect denoising process is removing the noise as much as possible on the premise of preserving the valuable signal. Although the DINF can remove the noise information of each input RoI feature by the adaptive thresholds, some useful signal information may also be removed by mistake since these thresholds cannot be exact. Therefore, similar to the structure of dense connection \cite{Dense-Connection}, in the iteration structure, we add all of the output RoI features of previous iterations together as the input of the next iteration. These dense connections help to transfer the signal information misfiltered by DINF in each iteration to the subsequent iterations. In INFN, the parameters of the DINF and the prediction heads (including the two fully-connection layers) are all shared in each iteration, hence multiple iterations will not increase the number of parameters. In INFN, every iteration will output a set of predictions (prediction boxes with classification). If $K$ iterations are conducted, including the original output of RCNN, there will be $K+1$ predictions in a single inference. The training targets are the same for all of the $K+1$ predictions during the training process.

\subsection{IoU Focal Factor}
In occluded pedestrian detection, we notice that the regression qualities of the overlapping objects are always worse than the non-overlapping objects. The bad regression quality boxes are incompact and ambiguous, making it difficult for NMS to judge which object they belong to. The focal loss was first proposed by \cite{Focal-Loss} to solve the imbalance issue between hard and easily classified examples in one-stage detectors. In occluded pedestrian detection, it is evident that the overlapping objects are hard-regressed samples compared with the non-overlapping objects since the features of heavily overlapping objects are too similar to be distinguished in the feature space. As shown in Fig2, there is also an imbalance among different overlapping degrees, i.e., the heavily overlapping samples are significantly less than lightly overlapping ones. Therefore, following the same principle of Focal Loss, we propose an IoU Focal Factor (IFF), which is used to modulate the regression loss function. It can be expressed as follows:
\begin{equation}
L_{IFF}=\mu \times L_{reg}
\end{equation}
\begin{equation}
\mu = -(1-IoU)^{\gamma}log(IoU)
\end{equation}
where IoU is the IoU between the prediction box and its corresponding ground-truth box. $\gamma \ge 0$ is the same focusing parameter as the focal loss. $L_{reg}$ is the box regression loss. In our paper, we adopt the SmoothL1 loss as the $L_{reg}$:
\begin{equation}
L_{reg}=\left\{
\begin{array}{cl}
0.5x^2, &  if \left |x \right | < 1 \\
\left|x \right |-0.5, &  otherwise \\
\end{array} \right.
\end{equation}

In the training process, even for the easily-regressed objects, the loss can hardly equal 0. The easily-regressed objects will still contribute to the loss in the training process even if their prediction boxes are already precise enough than other hard-regressed objects. Since the easily-regressed samples are the majority, their loss will cover the loss of minority hard-regressed samples. In IFF, we use the IoU between the prediction box and its corresponding ground truth box as the metric of regression quality. Using IoU, we produce a scaling parameter $\mu$, as shown in equation (5). When the IoU tends to be 1, meaning the corresponding box has been regressed to its target box precisely, $\mu$ will tend to be 0, and the corresponding loss will be significantly reduced. In contrast, when the IoU tends to be 0, $\mu$ tends to be 1, meaning that its corresponding loss will not be reduced sharply. Thus, the contribution of easily-regressed and hard-regressed objects to the final loss will be adjusted. When the influence of high-quality boxes is lessened, the influence of low-quality boxes will be highlighted naturally. 

\section{Experiment}
\subsection{Dataset and Settings}
In this paper, we adopt two pedestrian datasets, CrowdHuman and CityPersons\cite{CityPersons}. CrowdHuman is one of the most persuasive occluded pedestrian datasets. There are 22.64 objects per image, of which 2.40 objects are heavily overlapping (the IoU with other objects greater than 0.5).  The proportion of heavily overlapping objects in each image is about 10.6\%. CrowdHuman is separated into the training set, validation set, and testing set, which contain 15000, 4370, and 5000 images, respectively.   In this paper, we conduct experiments mainly on CrowdHuman, and all models are trained on the training set, and results are reported on the validation set. CityPersons is a common pedestrian detection benchmark. It contains 5000 images in total, separated as 2975 images for training, 500 images for validation, and 1525 images for testing, respectively.On average, there are 0.32 heavily overlapping objects per image in CityPersons. The proportion of heavily overlapping objects in each image is about 4.9\%. All related results are reported on the validation set.

Three metrics are adopted to evaluate the performance of the detector:
\begin{itemize}
\item AP (Average Precision) is the most popular metric for object detection, reflecting both the precision and recall ratios of the detection results. This paper reports the AP with the IoU threshold of 0.5, which is also written as $AP_{50}$ in other papers. Larger AP means better performance.
\item MR$^{-2}$ is the log-average miss rate on false positives per image in $[10^{-2}, 10^0]$. It is the most common metric in pedestrian detection, which is very sensitive to false positives (FPs). Improvement in FPs or recall ratios will both reduce the MR$^{-2}$. Smaller MR$^{-2}$ means better performance.
\item JI (Jaccard Index) evaluates how much the prediction set overlaps the ground truth set. In crowded object detection, it directly reflects the joint detection precision of a group of objects. Larger JI means better performance.
\end{itemize}
We adopt the Faster-RCNN with FPN\cite{FPN} as the baseline to conduct the ablation studies. Besides, to generate SOTA-comparable performance and proof that our DINF can be integrated into any RCNN-based detectors, we adopt the RCNN-based SOTA detector CrowdDet to compare with other current SOTA pedestrian detectors. The standard ResNet-50 pre-trained on ImgaeNet is used as the backbone network for both Faster-RCNN and CrowdDet. Moreover, we replace the original RoIPooling with the RoIAlign\cite{Mask-RCNN} for all detectors in this paper. The anchor scales are the same as FPN, while the aspect ratios are set to $H:W=\left\{1:1,2:1,3:1\right\}$. For training, the batch size is set to 16, split into 4 NVIDIA RTX3090 GPUs. Each training runs 40 epochs for both Faster-RCNN and CrowdDet. The ratio of positive to negative proposals for RoI is set to 1:1. For each image, the short edge is resized to 800 pixels during both training and testing. Finally, the NMS threshold for both Faster-RCNN and CrowdDet is set to 0.5. 

\subsection{Ablation Studies}
In this section, we conduct ablation studies on CrowdHuman. The Faster-RCNN with FPN is adopted as the baseline.
\paragraph{Ablation Studies on Proposed Methods} We first show the comprehensive results of our proposed DINF, INFN, and IFF, as shown in Table 1. The INFN$_{i}^{K}$ means the INFN is trained by $K$ iterations, and the results are obtained from the output of i-th iteration. We set $K=4$ in this section. 

\begin{table}[h]
  \centering
  \caption{Ablation Studies results on Proposed Methods}
    \begin{tabular}{rrrrr}
    \toprule
    \multicolumn{1}{l}{Model} & \multicolumn{1}{l}{$\gamma$} & \multicolumn{1}{l}{AP} & \multicolumn{1}{l}{MR} & \multicolumn{1}{l}{JI} \\
    \midrule
    \multicolumn{1}{l}{baseline} &   -   & 87.36 & 42.24 & 79.67 \\
    \multicolumn{1}{l}{+DINF} &        - & 87.67 & 41.75 & 79.98 \\
    \multicolumn{1}{l}{+INFN$_{1}^4$} &   -    & 89.08 & 41.58 & 81.45 \\
    \multicolumn{1}{l}{+INFN$_{3}^4$} &    -   & 88.42 & 41.08 & 80.72 \\
    \multicolumn{1}{l}{+INFN$_{1}^4$} & 0.1   & 89.21 & 41.09 & 81.28 \\
    \multicolumn{1}{l}{+INFN$_{3}^4$} & 0.1   & 88.59 & 40.98 & 80.69 \\
    \bottomrule
    \end{tabular}%
  \label{tab:addlabel}%
\end{table}%

The results in Table 1 show that the DINF without iteration structure increases AP and JI by 0.41\% and 0.31\% and reduces the MR$^{-2}$ by 0.49\%, respectively. When the DINF is integrated into the INFN, the improvements become more significant. From the testing results of the first iteration, AP and JI are increased by 1.72\% and 1.78\%, and the MR$^{-2}$ is reduced by 0.66\%, respectively. And from the results of the third iteration, the MR$^{-2}$ is reduced by 1.16\%, which is more pronounced compared with the first iteration, but the increments in AP and JI are relatively small, and they are increased by 1.06\% and 1.05\%. Thus, it can be seen that the best MR$^{-2}$ cannot be obtained in the iteration in which the AP and JI are the best, meaning we need to make a trade-off between the best AP and JI and the best MR$^{-2}$. However, the proposed IFF helps to bridge the MR$^{-2}$ gap between the first and third iteration. When the IFF with $\gamma=0.1$ is applied, compared with the baseline, the AP and JI are increased by 1.82\% and 1.61\%, respectively, and the MR$^{-2}$ is reduced by 1.15\%. With IFF applied, the improvement in MR$^{-2}$ of the first iteration is comparable to the third iteration, and the improvements in AP and JI are kept simultaneously. IFF makes the model optimization focus on the bad-regressed boxes and reduces the number of ambiguous boxes, which reduces false-positive predictions and thus improve the MR$^{-2}$. These above results demonstrate that our proposed methods effectively improve the detector's performance in occluded pedestrian detection.

\paragraph{Ablation studies on the iteration numbers of INFN} Our INFN can be trained in a multiple iterations manner, and the INIF trained by $K$ iterations will output $K+1$ predictions. Table 2 shows the results of the ablation studies on $K$, in which $i$ means the iteration number.The bold values indicate the best.

\begin{table}[htbp]
  \centering
  \caption{Ablation Studies results on iteration numbers on INFN}
   \renewcommand{\arraystretch}{0.7} 
    \begin{tabular}{ccccc}
    \toprule
    K     & i     & AP    & MR$^{-2}$    & JI \\
    \midrule
    baseline & - & 87.36 & 42.24 & 79.67 \\
    \midrule
    \multirow{2}[2]{*}{K=1} & 0     & 87.61 & 41.76 & 80.04 \\
          & 1     & 87.67 & 41.75 & 79.98 \\
    \midrule
    \multirow{3}[2]{*}{K=2} & 0     & 87.69 & 41.80  & 80.17 \\
          & 1     & 88.70  & 41.17 & 80.93 \\
          & 2     & 87.39 & 41.33 & 80.17 \\
    \midrule
    \multirow{4}[2]{*}{K=3} & 0     & 87.84 & 42.13 & 80.27 \\
          & 1     & \textbf{89.13} & 41.78 & 81.3 \\
          & 2     & 88.50  & 41.27 & 80.79 \\
          & 3     & 87.59 & 41.59 & 80.38 \\
    \midrule
    \multirow{5}[2]{*}{K=4} & 0     & 87.74 & 41.90  & 80.17 \\
          & 1     & 89.08 & 41.58 & \textbf{81.45} \\
          & 2     & 88.76 & 41.16 & 81.11 \\
          & 3     & 88.42 & \textbf{41.08} & 80.72 \\
          & 4     & 87.66 & 41.23 & 80.38 \\
    \midrule
    \multirow{6}[2]{*}{K=5} & 0     & 87.56 & 42.19 & 79.80 \\
          & 1     & 89.08 & 42.50  & 81.26 \\
          & 2     & 88.86 & 41.63 & 80.93 \\
          & 3     & 88.70  & 41.52 & 80.87 \\
          & 4     & 88.28 & 41.50  & 80.58 \\
          & 5     & 87.64 & 41.39 & 80.25 \\
    \midrule
    \multirow{7}[2]{*}{K=6} & 0     & 87.72 & 42.38 & 79.88 \\
          & 1     & 89.04 & 42.99 & 80.83 \\
          & 2     & 88.96 & 42.25 & 80.86 \\
          & 3     & 89.00   & 41.51 & 81.07 \\
          & 4     & 88.80  & 41.34 & 80.84 \\
          & 5     & 88.50  & 41.30  & 80.72 \\
          & 6     & 87.98 & 41.25 & 80.40 \\
    \bottomrule
    \end{tabular}%
  \label{tab:addlabel}%
\end{table}%

By analyzing the results from Table2, we obtain three properties of the iterable INFN. Firstly, the performance of the Faster-RCNN is significantly improved as $K$ increases. When $K=3$, our method achieves the best in AP, and AP is increased by 1.77\% compared with the baseline. When $K=4$, our method achieves the best in MR$^{-2}$ and JI. MR$^{-2}$ is improved by 1.16\%, and JI is improved by 1.78\%, respectively. Secondly, no matter how much $K$ is set, the first iteration always produces the best AP and JI. This property means that it is unnecessary to worry about much time consumption when we want the best performance in inference, i.e., we can train the model with $K>1$ and test the model with $K=1$ since the prediction head is shared in each iteration. Thirdly, $K$ is not  the bigger the better. When we set $K=5$ and $K=6$ in the training process, the AP and JI are not greater than that when $K=3$ or $K=4$. And the MR$^{-2}$ in the results of the first iteration of $K=5$ and $K=6$ are even worse than the baseline. All of the above descriptions can be shown intuitively in Fig6. It is worth noting that AP and JI of the output of the first iteration are the best among almost all of the iterations. And these two metrics will reduce in later iterations gradually as the iterations increase. However, the MR$^{-2}$ in the first iteration is not the best among all iterations. The iteration number of the best MR$^{-2}$ is not the same as it of the best of AP and JI, meaning that we cannot make the three metrics reach the best simultaneously in the same iteration. These results indicate that we need to make a trade-off between the numbers of iterations if the IFF is not applied. When we need high AP, JI, and inference speed, we can set the training $K$ to 3 or 4 and the testing $K$ to 1. When we need lower MR$^{-2}$, the testing K can be set to 3 or 4. Comprehensively, the $K=3$ or $K =4$ in the training process and $K=1$ in the testing process is a proper setting.

\begin{figure}[ht]%
\centering
\includegraphics[width=1.0\textwidth]{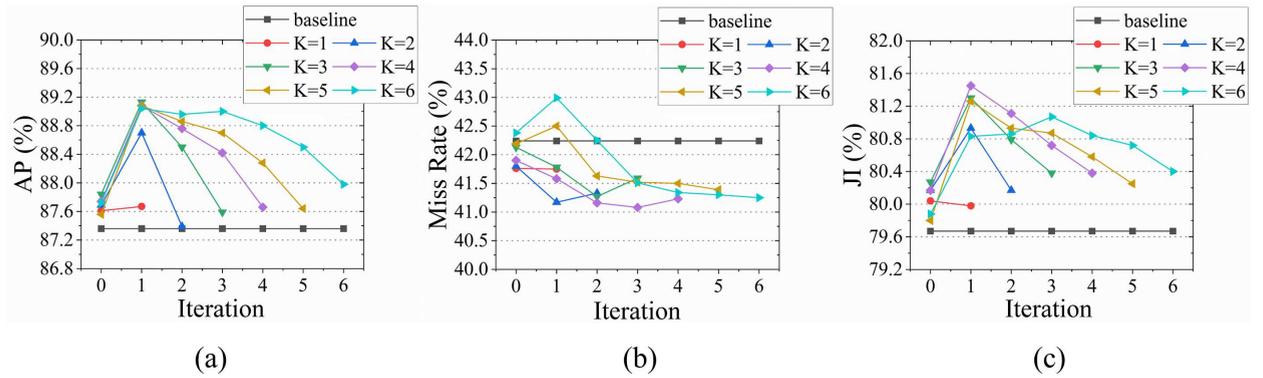}
\caption{Visualization of the testing results of INFN with different iteration numbers. (a) AP; (b) Miss Rate; (c) JI}\label{fig6}
\end{figure}

\paragraph{Ablation studies on the IFF} Similar to the focal loss, the $\gamma$ in IFF will also influence its performance. We conduct ablation studies on $\gamma$ with the original Faster-RCNN with FPN. Table 3 shows the ablation results. 

\begin{table}[htbp]
  \centering
  \caption{Ablation studies on the $\gamma$ of IFF}
    \begin{tabular}{lccc}
    \toprule
    \multicolumn{1}{l}{$\gamma$} & \multicolumn{1}{l}{AP} & \multicolumn{1}{l}{MR$^{-2}$} & \multicolumn{1}{l}{JI} \\
    \midrule
    \multicolumn{1}{l}{baseline} & 87.36 & 42.24 & 79.67 \\
    \midrule
    0.0   & 87.70  & \textbf{41.91} & \textbf{79.92} \\
    0.1   & 87.68 & 42.06 & 79.85 \\
    0.3   & 87.62  & 42.07 & 79.78 \\
    0.5   & \textbf{87.76} & 42.03 & 79.77 \\
    0.7   & 87.60  & 42.08 & 79.85 \\
    1.0   & 87.59 & 41.19 & 79.52 \\
    2.0   & 87.31 & 42.38 & 79.33 \\
    \bottomrule
    \end{tabular}%
  \label{tab:addlabel}%
\end{table}%
Compared with the baseline, IFF with $\gamma$=0.5 achieves the best in AP, and it is improved by 0.4\%. IFF with $\gamma$=0 achieves the best in MR$^{-2}$ and JI, and they are improved by 0.33\% and 0.25\%, respectively. In general, there is little difference in the three metrics when $\gamma$ is less than 1.0. We can still observe from the results that the performance gain gradually diminishes as $\gamma$ increases, and when $\gamma$ is greater than 1.0, the JI is less than the baseline, indicating that [0,1] is the appropriate interval for $\gamma$.

\subsection{Comparative Experiment}
In this section, we conduct comparative experiments on CrowdHuman and CityPersons with current SOTA pedestrian detectors. To generate competitive results and demonstrate our DINF can be integrated into any RCNN-based detectors, we apply our INFN to a SOTA RCNN-based detector CrowdDet, which proposed a set-NMS to overcome the shortcoming of traditional NMS in crowded scenes. 

\paragraph{CrowdHuman}
The current detectors can generally be divided into three categories, i.e., point-based, box-based, and query-based. Since the NMS process will significantly affect the final performance in crowded object detection, the query-based detectors benefit from the end-to-end advantage and have achieved a leading performance. To reduce the interference of NMS on the performance, we applied the INFN to the SOTA box-based detector CrowdDet equipped with a set-NMS. Besides, we also applied the soft-NMS to Faster-RCNN to make up for its shortcomings in terms of NMS. The comparison results with other SOTA pedestrian detectors are shown in Table4, in which the $IFF_j$ means the IFF with $\gamma=j$. 

\begin{table}[htbp]
  \centering
  \begin{threeparttable}[b]
  \caption{Comparison Results with SOTA methods on CrowdHuman}
\renewcommand{\arraystretch}{0.7}
    \begin{tabular}{lcrrr}
    \toprule
    Model & \multicolumn{1}{c}{Type} & \multicolumn{1}{c}{AP} & \multicolumn{1}{c}{MR} & \multicolumn{1}{c}{JI} \\
    \midrule
    FCOS\cite{FCOS}  & \multirow{3}[2]{*}{point-based} & 86.80  & 54.00    & 75.70 \\
    FCOS\cite{FCOS}+MIP\cite{CrowdDet} &       & 87.30  & 51.20  & 77.30 \\
    POTO\cite{POTO}  &       & 89.10  & 47.80  & 79.30 \\
    \midrule
    RetinaNet\cite{CrowdHuman} &   \multirow{10}[2]{*}{box-based}    & 80.83 & 63.33 & \multicolumn{1}{l}{-} \\
    Adaptive NMS\cite{Adaptive-NMS} &       & 84.71 & 49.73 & \multicolumn{1}{l}{-} \\
    FPN\cite{CrowdHuman}   &       & 85.60  & 55.94 & \multicolumn{1}{l}{-} \\
    Repulsion Loss\cite{Repulsion-Loss} &       & 85.64 & 45.69 & \multicolumn{1}{l}{-} \\
    ATSS\cite{ATSS}  &       & 87.00    & 51.10  & 75.90 \\
    ATSS\cite{ATSS}+MIP\cite{CrowdDet} &       & 88.70  & 51.60  & 77.00 \\
    FPN\cite{CrowdDet}   &       & 85.80  & 42.90  & 79.80 \\
    PBM\cite{PBM}   &       & 89.29 & 43.35 & \multicolumn{1}{l}{-} \\
    Beta-RCNN\cite{Beta-RCNN} &       & 89.60  & 40.60  & \multicolumn{1}{l}{-} \\
    CrowdDet\cite{CrowdDet} &       & 90.70  & 41.40  & 82.30 \\
    \midrule
    DETR\cite{DETR}  & \multirow{5}[2]{*}{query-based} & 75.90  & 73.20  & 74.40 \\
    PEDR\cite{PEDR}  &       & 91.60  & 43.70  & \textbf{83.30} \\
    D-DETR\cite{D-DETR} &       & 91.50  & 43.70  & 83.10 \\
    S-RCNN\cite{Sparse-RCNN} &       & 91.30  & 44.80  & 81.30 \\
    Iter-E2EDET\cite{Iter-E2EDET} &       & \textbf{92.50}  & 41.60  & \textbf{83.30} \\
    \midrule
    Baseline & \multirow{6}[2]{*}{\makecell[c]{box-based\\(\emph{Our imple.})}} & 87.36 & 42.24 & 79.67 \\
    Baseline+INFN$_1^4$+IFF$_{0.1}$ &       & 89.21 & 41.09 & 81.28 \\
    Baseline+INFN$_1^4$+IFF$_{0.1}$(soft-NMS\cite{Soft-NMS}) &       & 90.84 & 41.08 & 81.26 \\
    CrowdDet &       & 90.72 & 41.13 & 82.60 \\
    CrowdDet+INFN$_1^3$ &       & 90.85 & \textbf{40.35} & 82.82 \\
    \bottomrule
    \end{tabular}%
    \begin{tablenotes}
     \item[1] Baseline is Faster-RCNN with FPN
     \item[2] \emph{our imple.} represents the results of the experiments we implemented, and the same goes for the following.
    \end{tablenotes}
    \end{threeparttable}
  \label{tab:addlabel}%
\end{table}%

As shown in Table4, the Faster-RCNN with soft-NMS achieves a SOTA performance in AP among all box-based detectors. For CrowdDet, our INFN helps it achieve the best performance compared with all of the current box-based detectors. The INFN achieves 0.13\% AP, 0.78\% MR$^{-2}$, and 0.22\% JI gains over the original CrowdDet. It also achieves 1.25\% AP and 0.25\% MR$^{-2}$ gains over the SOTA box-based detector Beta-RCNN. As for the query-based methods, benefiting from their end-to-end framework, they achieve a relatively higher recall rate, which contributes much to the AP but increases the false-positive rate. The false-positive will significantly impact the MR$^{-2}$, which is the main metric in pedestrian detection and reflects the balance between false-positive and recall rate. Although the performance of CrowdDet with INFN is slightly poorer than the query-based methods in AP and JI, our method has a big lead in MR$^{-2}$ and achieves 1.25\% MR$^{-2}$ gains over the best query-based detector Iter-E2EDet specifically. The SOTA performance demonstrates the effectiveness of our approaches.

\paragraph{CityPersons} is one of the wildly used pedestrian detection benchmarks. Each image has a size of $1024\times2048$. Following the mainstream method for better performance, we train and evaluate our models with the image resolution enlarged by 1.3X compared with the original ones. All models are trained on the reasonable subset(height$>50$, occlusion$<0.35$). Following \cite{Adaptive-NMS}, we adopt AggLoss\cite{Occlusion-Aware} for a strong baseline. Other training configurations are the same as \cite{CityPersons}. Table5 shows the results, in which the \textbf{HO} means the Miss Rate on heavily overlapping subset(occlusion $\in [0.2,0.65]$) and it reflects detectors' ability on occlusion detection. \textbf{R} means the Miss Rate on reasonable subset, reflecting the comparhensive performance.
\begin{table}[htp]
  \centering
  \caption{Comparison results with SOTA methods on CityPersons}
\renewcommand{\arraystretch}{0.7}
    \begin{tabular}{lccc}
    \toprule
    Model & Backbone & \textbf{R}     & \textbf{HO} \\
    \midrule
    Repulsion Loss\cite{Repulsion-Loss} & ResNet-50 &   11.57    & 54.78 \\
    OR-CNN\cite{Occlusion-Aware} & VGG-16 & 11.00    & 51.90 \\
    Bi-box\cite{Bi-box} & VGG-16+ & 11.53 & - \\
    Adaptive NMS\cite{Adaptive-NMS} & VGG-16 & 10.80  & 54.00 \\
    Faster-RCNN\cite{Guided-Attention} & ResNet-50 & 15.52 & 54.74 \\
    ATT-Part\cite{Guided-Attention} & ResNet-50 & 15.96 &\textbf{46.52} \\
    ATT-self\cite{Guided-Attention} & ResNet-50 & 20.93 & 51.21 \\
    NOH-NMS\cite{NOH-NMS} & ResNet-50 & 10.80  & 53.00 \\
    CrowdDet\cite{CrowdDet} & ResNet-50 & 10.70  & - \\
    \midrule
    Faster-RCNN(\emph{our imple.}) & ResNet-50 & 11.68 & 53.87 \\
    Faster-RCNN+INFN$_1^3$+IFF$_{0.1}$ & ResNet-50 & 11.31 & 52.50 \\
    CrowdDet(\emph{our imple.}) & ResNet-50 & 10.68 & 53.42 \\
    CrowdDet+INFN$_1^3$ & ResNet-50 & \textbf{10.32} & 52.25 \\
    \bottomrule
    \end{tabular}%
  \label{tab:addlabel}%
\end{table}%

The results in Table5 show that our methods can help the Faster-RCNN achieve 0.31\% gains in \textbf{R} and 1.37\% gains in \textbf{HO}. The significant improvements in \textbf{HO} indicate that our INFN is more effective on the occluded object detection, benefitting from the noise filtering function of INFN. Futhermore, the CrowdDet with INFN achieves the best performance among all listed methods. Compared with the results on CrowdHuman, the enhancements on CityPersons are less significant. We speculate that because the proportion of heavily overlapping objects in each image in Citypersons (4.9\%) is less it that in CrowdHuman (10.6\%).And in the \emph{Reasonable} subset of CityPersons, the proportion of heavily overlapping objects will be lower further.  Since our methods are mainly used to solve the occlusion issue, more heavily overlapping objects are more indicative of the effectiveness of our proposed methods. Above results still demonstrate that our INFN can further enhance the performance of RCNN-based pedestrian detectors, especially in terms of occlusion detection.
\subsection{Disscussion}
In order to observe the influence of DINF on the feature vector of the instance more intuitively, we randomly select two images and implement dimensionality reduction to the output instance features using the principal component analysis (PCA). For ease of visualization, the dimension is reduced from 1024 to 2. We compare the instance features generated by Faster RCNN with/without INFN (K=3). The results are shown in Fig7. 
\begin{figure}[ht]%
\centering
\includegraphics[width=1.0\textwidth]{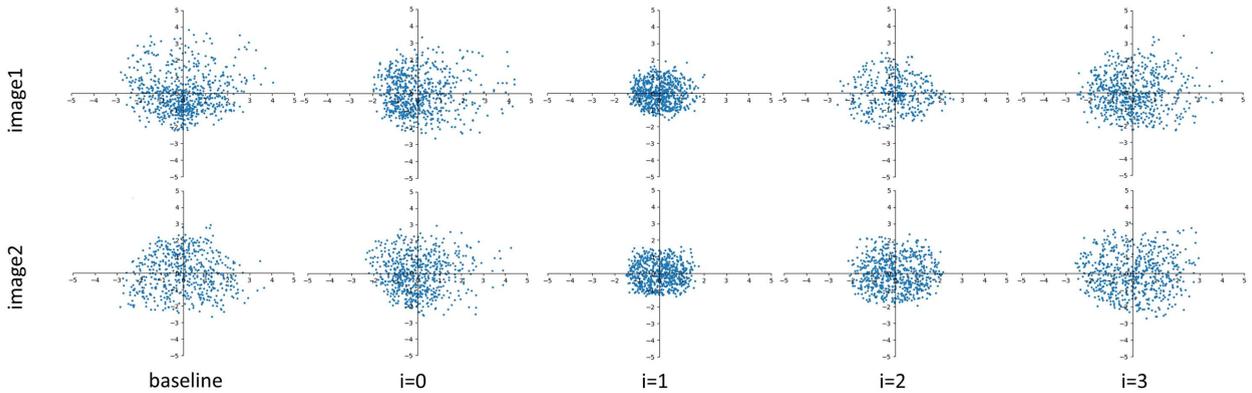}
\caption{Visualization of the dimension-reduced instance feature}\label{fig7}
\end{figure}
As shown in Fig7, the degree of compactness of the dimension-reduced feature exhibits a strong correlation with the detectors’ performance. As shown in Fig6, when $i$=1, the corresponding instance features are the most compact, and their corresponding performance is the best as well. With the increase of $i$, the performance is declining, and the compactness of instance features generally also reduces. In Fig7, the compactness reflects the variance of instance features. And the small variance of the predictions is always craved in learning systems. Small variance means robustness and anti-interference ability, which explains why the best performance is obtained when $i$=1. 

In addition, the above experimental results show that compared to the single-shot inference of DIFN, the iterative structure of INFN effectively boosts the performance. In INFN, the parameters of prediction heads in each iteration are shared, meaning that the improvements are not benefitting from extra network structures. The only difference between the single-shot and the iterative modes of DINF is that in each training iteration, the prediction heads in INFN are trained by multiple RoI features with different SNR levels, which can be regarded as a data augmentation at the feature level. In the one-shot training mode, each instance only updates the parameters of the prediction head once in each training iteration. However, when the INFN with $K>1$ is applied, the prediction head parameters can be updated multiple times by each instance in each training iteration. Due to the adaptively filtering function of the DINF, the input instance features of the prediction head in each iteration are of various SNR levels, enhancing the generalization ability of the shared prediction heads to instance features with different SNR levels. The enhanced generalization ability makes the prediction head more robust to the input instance features with various SNR levels, which can be demonstrated from Fig6 that when the INFN with $K>1$ is applied, the results generated by the original RoI features ($i$=0) still outperform the baseline. Or, we can find in Fig7 that the features are more compact when $i$=0 compared to the baseline, even though they are absolutely the same in terms of the input features and network structures.

\subsection{Qualitative results}
\begin{figure}[ht]%
\centering
\includegraphics[width=1.0\textwidth]{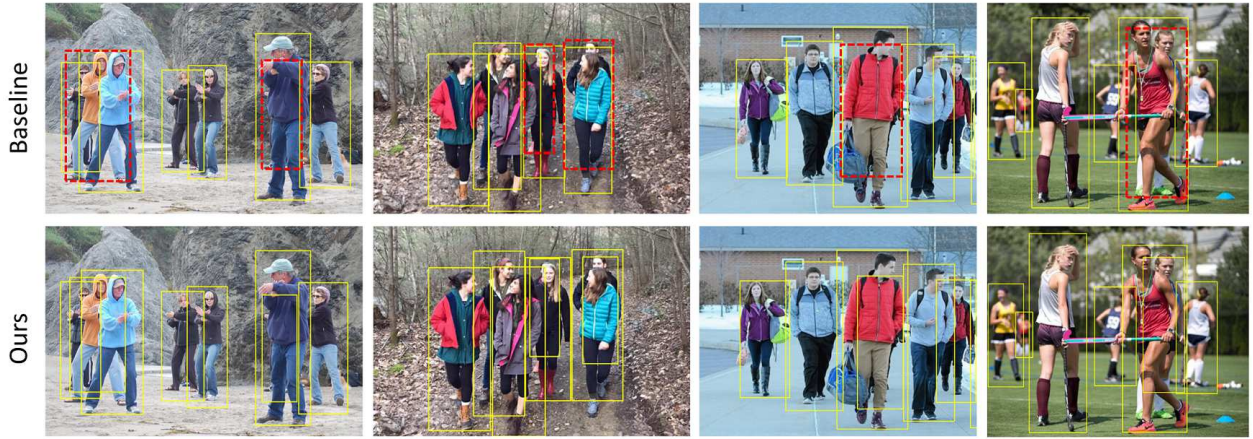}
\caption{Qualitative results comparison between proposed methods and baseline. The first and second rows show the results generated by the baseline and our proposed methods, respectively. The red dotted boxes in the first row indicate the heavily occluded objects missed by the baseline detector}\label{fig8}
\end{figure}

In this section, we show the effectiveness of our proposed methods on occluded objects detection by the qualitative results, as shown in Fig8. The first row is the result generated by the baseline detector, and the the second row is the result genarated by our proposed method. Compared with the first row, some heavily occluded objects which are even hard to be seen are detected by our proposed methods. As shown by the red doted boxes, some of these heavily occluded objects have very limited visible pixels, so their information are easily covered up by the occluders. With the noise filter function of our DINF, these heavily occluded objects are detected successfully.

\section{Conclusion}
In this paper, we consider occlusion issue of pedestrian detection from the perspective of denoising and propose a dynamic instance noise filter DINF that simulates the wavelet denoising process. To enhance the performance of DINF, an iterative noise filter network INFN is proposed, which uses DINF to filter noise multiple times in an iterative manner to boost noise filtering ability. The INFN implements an instance-feature-level data augmentation, which improves the generalization ability and robustness of the prediction head. In addition, the imbalance issue between easily-regressed and hard-regressed samples is proposed. To solve this problem, an IoU focal factor IFF is proposed to down-weights the loss assigned to well-regressed boxes. Extensive experiments are conducted to show that our proposed methods can achieve state-of-the-art performance.



\bibliographystyle{cas-model2-names}
\bibliography{cas-refs}

\begin{thebibliography}{55}
\expandafter\ifx\csname natexlab\endcsname\relax\def\natexlab#1{#1}\fi
\providecommand{\url}[1]{\texttt{#1}}
\providecommand{\href}[2]{#2}
\providecommand{\path}[1]{#1}
\providecommand{\DOIprefix}{doi:}
\providecommand{\ArXivprefix}{arXiv:}
\providecommand{\URLprefix}{URL: }
\providecommand{\Pubmedprefix}{pmid:}
\providecommand{\doi}[1]{\href{http://dx.doi.org/#1}{\path{#1}}}
\providecommand{\Pubmed}[1]{\href{pmid:#1}{\path{#1}}}
\providecommand{\bibinfo}[2]{#2}
\ifx\xfnm\relax \def\xfnm[#1]{\unskip,\space#1}\fi
\bibitem[{Bochkovskiy et~al.(2020)Bochkovskiy, Wang and Liao}]{YOLOv4}
\bibinfo{author}{Bochkovskiy, A.}, \bibinfo{author}{Wang, C.},
  \bibinfo{author}{Liao, H.M.}, \bibinfo{year}{2020}.
\newblock \bibinfo{title}{Yolov4: Optimal speed and accuracy of object
  detection}.
\newblock \bibinfo{journal}{CoRR} \bibinfo{volume}{abs/2004.10934}.
\newblock \URLprefix \url{https://arxiv.org/abs/2004.10934},
  \href{http://arxiv.org/abs/2004.10934}{\tt arXiv:2004.10934}.
\bibitem[{Bodla et~al.(2017)Bodla, Singh, Chellappa and Davis}]{Soft-NMS}
\bibinfo{author}{Bodla, N.}, \bibinfo{author}{Singh, B.},
  \bibinfo{author}{Chellappa, R.}, \bibinfo{author}{Davis, L.S.},
  \bibinfo{year}{2017}.
\newblock \bibinfo{title}{Soft-nms 〞 improving object detection with one line
  of code}, in: \bibinfo{booktitle}{2017 IEEE International Conference on
  Computer Vision (ICCV)}, pp. \bibinfo{pages}{5562--5570}.
\newblock \DOIprefix\doi{10.1109/ICCV.2017.593}.
\bibitem[{Carion et~al.(2020)Carion, Massa, Synnaeve, Usunier, Kirillov and
  Zagoruyko}]{DETR}
\bibinfo{author}{Carion, N.}, \bibinfo{author}{Massa, F.},
  \bibinfo{author}{Synnaeve, G.}, \bibinfo{author}{Usunier, N.},
  \bibinfo{author}{Kirillov, A.}, \bibinfo{author}{Zagoruyko, S.},
  \bibinfo{year}{2020}.
\newblock \bibinfo{title}{End-to-end object detection with transformers}, in:
  \bibinfo{editor}{Vedaldi, A.}, \bibinfo{editor}{Bischof, H.},
  \bibinfo{editor}{Brox, T.}, \bibinfo{editor}{Frahm, J.M.} (Eds.),
  \bibinfo{booktitle}{Computer Vision -- ECCV 2020},
  \bibinfo{publisher}{Springer International Publishing},
  \bibinfo{address}{Cham}. pp. \bibinfo{pages}{213--229}.
\bibitem[{Chu et~al.(2020)Chu, Zheng, Zhang and Sun}]{CrowdDet}
\bibinfo{author}{Chu, X.}, \bibinfo{author}{Zheng, A.}, \bibinfo{author}{Zhang,
  X.}, \bibinfo{author}{Sun, J.}, \bibinfo{year}{2020}.
\newblock \bibinfo{title}{Detection in crowded scenes: One proposal, multiple
  predictions}, in: \bibinfo{booktitle}{2020 IEEE/CVF Conference on Computer
  Vision and Pattern Recognition (CVPR)}, pp. \bibinfo{pages}{12211--12220}.
\newblock \DOIprefix\doi{10.1109/CVPR42600.2020.01223}.
\bibitem[{Dai et~al.(2016a)Dai, He and Sun}]{RoI-Warp}
\bibinfo{author}{Dai, J.}, \bibinfo{author}{He, K.}, \bibinfo{author}{Sun, J.},
  \bibinfo{year}{2016}a.
\newblock \bibinfo{title}{Instance-aware semantic segmentation via multi-task
  network cascades}, in: \bibinfo{booktitle}{2016 IEEE Conference on Computer
  Vision and Pattern Recognition (CVPR)}, pp. \bibinfo{pages}{3150--3158}.
\newblock \DOIprefix\doi{10.1109/CVPR.2016.343}.
\bibitem[{Dai et~al.(2016b)Dai, Li, He and Sun}]{PS-RoI-Pooling}
\bibinfo{author}{Dai, J.}, \bibinfo{author}{Li, Y.}, \bibinfo{author}{He, K.},
  \bibinfo{author}{Sun, J.}, \bibinfo{year}{2016}b.
\newblock \bibinfo{title}{R-fcn: Object detection via region-based fully
  convolutional networks}, in: \bibinfo{booktitle}{Proceedings of the 30th
  International Conference on Neural Information Processing Systems},
  \bibinfo{publisher}{Curran Associates Inc.}, \bibinfo{address}{Red Hook, NY,
  USA}. p. \bibinfo{pages}{379每387}.
\bibitem[{De~Brabandere et~al.(2016)De~Brabandere, Jia, Tuytelaars and
  Van~Gool}]{Dynamic-filter-networks}
\bibinfo{author}{De~Brabandere, B.}, \bibinfo{author}{Jia, X.},
  \bibinfo{author}{Tuytelaars, T.}, \bibinfo{author}{Van~Gool, L.},
  \bibinfo{year}{2016}.
\newblock \bibinfo{title}{Dynamic filter networks}, in:
  \bibinfo{booktitle}{Proceedings of the 30th International Conference on
  Neural Information Processing Systems}, \bibinfo{publisher}{Curran Associates
  Inc.}, \bibinfo{address}{Red Hook, NY, USA}. p. \bibinfo{pages}{667每675}.
\bibitem[{Donoho(1995)}]{De-noising-by-soft-thresholding}
\bibinfo{author}{Donoho, D.}, \bibinfo{year}{1995}.
\newblock \bibinfo{title}{De-noising by soft-thresholding}.
\newblock \bibinfo{journal}{IEEE Transactions on Information Theory}
  \bibinfo{volume}{41}, \bibinfo{pages}{613--627}.
\newblock \DOIprefix\doi{10.1109/18.382009}.
\bibitem[{Duan et~al.(2010)Duan, Ai and
  Lao}]{A-Structural-Filter-Approach-to-Human}
\bibinfo{author}{Duan, G.}, \bibinfo{author}{Ai, H.}, \bibinfo{author}{Lao,
  S.}, \bibinfo{year}{2010}.
\newblock \bibinfo{title}{A structural filter approach to human detection}, in:
  \bibinfo{booktitle}{Proceedings of the 11th European Conference on Computer
  Vision: Part VI}, \bibinfo{publisher}{Springer-Verlag},
  \bibinfo{address}{Berlin, Heidelberg}. p. \bibinfo{pages}{238每251}.
\bibitem[{Fu et~al.(2017)Fu, Liu, Ranga, Tyagi and Berg}]{DSSD}
\bibinfo{author}{Fu, C.}, \bibinfo{author}{Liu, W.}, \bibinfo{author}{Ranga,
  A.}, \bibinfo{author}{Tyagi, A.}, \bibinfo{author}{Berg, A.C.},
  \bibinfo{year}{2017}.
\newblock \bibinfo{title}{{DSSD} : Deconvolutional single shot detector}.
\newblock \bibinfo{journal}{CoRR} \bibinfo{volume}{abs/1701.06659}.
\newblock \URLprefix \url{http://arxiv.org/abs/1701.06659},
  \href{http://arxiv.org/abs/1701.06659}{\tt arXiv:1701.06659}.
\bibitem[{G{\"{a}}hlert et~al.(2020)G{\"{a}}hlert, Hanselmann, Franke and
  Denzler}]{Visibility-Guided-NMS}
\bibinfo{author}{G{\"{a}}hlert, N.}, \bibinfo{author}{Hanselmann, N.},
  \bibinfo{author}{Franke, U.}, \bibinfo{author}{Denzler, J.},
  \bibinfo{year}{2020}.
\newblock \bibinfo{title}{Visibility guided {NMS:} efficient boosting of amodal
  object detection in crowded traffic scenes}.
\newblock \bibinfo{journal}{CoRR} \bibinfo{volume}{abs/2006.08547}.
\newblock \URLprefix \url{https://arxiv.org/abs/2006.08547},
  \href{http://arxiv.org/abs/2006.08547}{\tt arXiv:2006.08547}.
\bibitem[{Girshick(2015)}]{Fast-RCNN}
\bibinfo{author}{Girshick, R.}, \bibinfo{year}{2015}.
\newblock \bibinfo{title}{Fast r-cnn}, in: \bibinfo{booktitle}{2015 IEEE
  International Conference on Computer Vision (ICCV)}, pp.
  \bibinfo{pages}{1440--1448}.
\newblock \DOIprefix\doi{10.1109/ICCV.2015.169}.
\bibitem[{Girshick et~al.(2014)Girshick, Donahue, Darrell and Malik}]{RCNN}
\bibinfo{author}{Girshick, R.}, \bibinfo{author}{Donahue, J.},
  \bibinfo{author}{Darrell, T.}, \bibinfo{author}{Malik, J.},
  \bibinfo{year}{2014}.
\newblock \bibinfo{title}{Rich feature hierarchies for accurate object
  detection and semantic segmentation}, in: \bibinfo{booktitle}{2014 IEEE
  Conference on Computer Vision and Pattern Recognition}, pp.
  \bibinfo{pages}{580--587}.
\newblock \DOIprefix\doi{10.1109/CVPR.2014.81}.
\bibitem[{He et~al.(2017a)He, Gkioxari, Doll芍r and Girshick}]{RoI-Align}
\bibinfo{author}{He, K.}, \bibinfo{author}{Gkioxari, G.},
  \bibinfo{author}{Doll芍r, P.}, \bibinfo{author}{Girshick, R.},
  \bibinfo{year}{2017}a.
\newblock \bibinfo{title}{Mask r-cnn}, in: \bibinfo{booktitle}{2017 IEEE
  International Conference on Computer Vision (ICCV)}, pp.
  \bibinfo{pages}{2980--2988}.
\newblock \DOIprefix\doi{10.1109/ICCV.2017.322}.
\bibitem[{He et~al.(2017b)He, Gkioxari, Doll芍r and Girshick}]{Mask-RCNN}
\bibinfo{author}{He, K.}, \bibinfo{author}{Gkioxari, G.},
  \bibinfo{author}{Doll芍r, P.}, \bibinfo{author}{Girshick, R.},
  \bibinfo{year}{2017}b.
\newblock \bibinfo{title}{Mask r-cnn}, in: \bibinfo{booktitle}{2017 IEEE
  International Conference on Computer Vision (ICCV)}, pp.
  \bibinfo{pages}{2980--2988}.
\newblock \DOIprefix\doi{10.1109/ICCV.2017.322}.
\bibitem[{Hu et~al.(2020)Hu, Shen, Albanie, Sun and Wu}]{SENet}
\bibinfo{author}{Hu, J.}, \bibinfo{author}{Shen, L.}, \bibinfo{author}{Albanie,
  S.}, \bibinfo{author}{Sun, G.}, \bibinfo{author}{Wu, E.},
  \bibinfo{year}{2020}.
\newblock \bibinfo{title}{Squeeze-and-excitation networks}.
\newblock \bibinfo{journal}{IEEE Transactions on Pattern Analysis and Machine
  Intelligence} \bibinfo{volume}{42}, \bibinfo{pages}{2011--2023}.
\newblock \DOIprefix\doi{10.1109/TPAMI.2019.2913372}.
\bibitem[{Huang et~al.(2017)Huang, Liu, Van Der~Maaten and
  Weinberger}]{Dense-Connection}
\bibinfo{author}{Huang, G.}, \bibinfo{author}{Liu, Z.}, \bibinfo{author}{Van
  Der~Maaten, L.}, \bibinfo{author}{Weinberger, K.Q.}, \bibinfo{year}{2017}.
\newblock \bibinfo{title}{Densely connected convolutional networks}, in:
  \bibinfo{booktitle}{2017 IEEE Conference on Computer Vision and Pattern
  Recognition (CVPR)}, pp. \bibinfo{pages}{2261--2269}.
\newblock \DOIprefix\doi{10.1109/CVPR.2017.243}.
\bibitem[{Huang et~al.(2020)Huang, Ge, Jie and Yoshie}]{PBM}
\bibinfo{author}{Huang, X.}, \bibinfo{author}{Ge, Z.}, \bibinfo{author}{Jie,
  Z.}, \bibinfo{author}{Yoshie, O.}, \bibinfo{year}{2020}.
\newblock \bibinfo{title}{Nms by representative region: Towards crowded
  pedestrian detection by proposal pairing}, in: \bibinfo{booktitle}{2020
  IEEE/CVF Conference on Computer Vision and Pattern Recognition (CVPR)}, pp.
  \bibinfo{pages}{10747--10756}.
\newblock \DOIprefix\doi{10.1109/CVPR42600.2020.01076}.
\bibitem[{Jiang et~al.(2018)Jiang, Luo, Mao, Xiao and
  Jiang}]{Precise-RoI-Pooling}
\bibinfo{author}{Jiang, B.}, \bibinfo{author}{Luo, R.}, \bibinfo{author}{Mao,
  J.}, \bibinfo{author}{Xiao, T.}, \bibinfo{author}{Jiang, Y.},
  \bibinfo{year}{2018}.
\newblock \bibinfo{title}{Acquisition of localization confidence for accurate
  object detection}.
\newblock \bibinfo{journal}{CoRR} \bibinfo{volume}{abs/1807.11590}.
\newblock \URLprefix \url{http://arxiv.org/abs/1807.11590},
  \href{http://arxiv.org/abs/1807.11590}{\tt arXiv:1807.11590}.
\bibitem[{Law and Deng(2020)}]{Cornernet}
\bibinfo{author}{Law, H.}, \bibinfo{author}{Deng, J.}, \bibinfo{year}{2020}.
\newblock \bibinfo{title}{Cornernet: Detecting objects as paired keypoints}.
\newblock \bibinfo{journal}{International Journal of Computer Vision}
  \bibinfo{volume}{128}, \bibinfo{pages}{642--656}.
\newblock \DOIprefix\doi{10.1007/s11263-019-01204-1}.
\bibitem[{Lin et~al.(2020a)Lin, Li, Bu, Sun, Lin, Yan, Ouyang and Deng}]{PEDR}
\bibinfo{author}{Lin, M.}, \bibinfo{author}{Li, C.}, \bibinfo{author}{Bu, X.},
  \bibinfo{author}{Sun, M.}, \bibinfo{author}{Lin, C.}, \bibinfo{author}{Yan,
  J.}, \bibinfo{author}{Ouyang, W.}, \bibinfo{author}{Deng, Z.},
  \bibinfo{year}{2020}a.
\newblock \bibinfo{title}{{DETR} for pedestrian detection}.
\newblock \bibinfo{journal}{CoRR} \bibinfo{volume}{abs/2012.06785}.
\newblock \URLprefix \url{https://arxiv.org/abs/2012.06785},
  \href{http://arxiv.org/abs/2012.06785}{\tt arXiv:2012.06785}.
\bibitem[{Lin et~al.(2017)Lin, Doll芍r, Girshick, He, Hariharan and
  Belongie}]{FPN}
\bibinfo{author}{Lin, T.Y.}, \bibinfo{author}{Doll芍r, P.},
  \bibinfo{author}{Girshick, R.}, \bibinfo{author}{He, K.},
  \bibinfo{author}{Hariharan, B.}, \bibinfo{author}{Belongie, S.},
  \bibinfo{year}{2017}.
\newblock \bibinfo{title}{Feature pyramid networks for object detection}, in:
  \bibinfo{booktitle}{2017 IEEE Conference on Computer Vision and Pattern
  Recognition (CVPR)}, pp. \bibinfo{pages}{936--944}.
\newblock \DOIprefix\doi{10.1109/CVPR.2017.106}.
\bibitem[{Lin et~al.(2020b)Lin, Goyal, Girshick, He and Doll芍r}]{Focal-Loss}
\bibinfo{author}{Lin, T.Y.}, \bibinfo{author}{Goyal, P.},
  \bibinfo{author}{Girshick, R.}, \bibinfo{author}{He, K.},
  \bibinfo{author}{Doll芍r, P.}, \bibinfo{year}{2020}b.
\newblock \bibinfo{title}{Focal loss for dense object detection}.
\newblock \bibinfo{journal}{IEEE Transactions on Pattern Analysis and Machine
  Intelligence} \bibinfo{volume}{42}, \bibinfo{pages}{318--327}.
\newblock \DOIprefix\doi{10.1109/TPAMI.2018.2858826}.
\bibitem[{Liu et~al.(2019)Liu, Huang and Wang}]{Adaptive-NMS}
\bibinfo{author}{Liu, S.}, \bibinfo{author}{Huang, D.}, \bibinfo{author}{Wang,
  Y.}, \bibinfo{year}{2019}.
\newblock \bibinfo{title}{Adaptive nms: Refining pedestrian detection in a
  crowd}, in: \bibinfo{booktitle}{2019 IEEE/CVF Conference on Computer Vision
  and Pattern Recognition (CVPR)}, pp. \bibinfo{pages}{6452--6461}.
\newblock \DOIprefix\doi{10.1109/CVPR.2019.00662}.
\bibitem[{Liu et~al.(2016)Liu, Anguelov, Erhan, Szegedy, Reed, Fu and
  Berg}]{SSD}
\bibinfo{author}{Liu, W.}, \bibinfo{author}{Anguelov, D.},
  \bibinfo{author}{Erhan, D.}, \bibinfo{author}{Szegedy, C.},
  \bibinfo{author}{Reed, S.}, \bibinfo{author}{Fu, C.Y.},
  \bibinfo{author}{Berg, A.C.}, \bibinfo{year}{2016}.
\newblock \bibinfo{title}{Ssd: Single shot multibox detector}, in:
  \bibinfo{booktitle}{Computer Vision -- ECCV 2016},
  \bibinfo{publisher}{Springer International Publishing},
  \bibinfo{address}{Cham}. pp. \bibinfo{pages}{21--37}.
\bibitem[{Mathias et~al.(2013)Mathias, Benenson, Timofte and
  Gool}]{Handling-Occlusions-with-Franken-Classifiers}
\bibinfo{author}{Mathias, M.}, \bibinfo{author}{Benenson, R.},
  \bibinfo{author}{Timofte, R.}, \bibinfo{author}{Gool, L.V.},
  \bibinfo{year}{2013}.
\newblock \bibinfo{title}{Handling occlusions with franken-classifiers}, in:
  \bibinfo{booktitle}{2013 IEEE International Conference on Computer Vision},
  pp. \bibinfo{pages}{1505--1512}.
\newblock \DOIprefix\doi{10.1109/ICCV.2013.190}.
\bibitem[{Ouyang and
  Wang(2012)}]{A-discriminative-deep-model-for-pedestrian-detection-with-occlusion-handling}
\bibinfo{author}{Ouyang, W.}, \bibinfo{author}{Wang, X.}, \bibinfo{year}{2012}.
\newblock \bibinfo{title}{A discriminative deep model for pedestrian detection
  with occlusion handling}, in: \bibinfo{booktitle}{2012 IEEE Conference on
  Computer Vision and Pattern Recognition}, pp. \bibinfo{pages}{3258--3265}.
\newblock \DOIprefix\doi{10.1109/CVPR.2012.6248062}.
\bibitem[{Ouyang and Wang(2013)}]{Joint-Deep-Learning-for-Pedestrian-Detection}
\bibinfo{author}{Ouyang, W.}, \bibinfo{author}{Wang, X.}, \bibinfo{year}{2013}.
\newblock \bibinfo{title}{Joint deep learning for pedestrian detection}, in:
  \bibinfo{booktitle}{2013 IEEE International Conference on Computer Vision},
  pp. \bibinfo{pages}{2056--2063}.
\newblock \DOIprefix\doi{10.1109/ICCV.2013.257}.
\bibitem[{Redmon et~al.(2016)Redmon, Divvala, Girshick and Farhadi}]{YOLOv1}
\bibinfo{author}{Redmon, J.}, \bibinfo{author}{Divvala, S.},
  \bibinfo{author}{Girshick, R.}, \bibinfo{author}{Farhadi, A.},
  \bibinfo{year}{2016}.
\newblock \bibinfo{title}{You only look once: Unified, real-time object
  detection}, in: \bibinfo{booktitle}{Proceedings of the IEEE Conference on
  Computer Vision and Pattern Recognition (CVPR)}.
\bibitem[{Redmon and Farhadi(2017)}]{YOLOv2}
\bibinfo{author}{Redmon, J.}, \bibinfo{author}{Farhadi, A.},
  \bibinfo{year}{2017}.
\newblock \bibinfo{title}{Yolo9000: Better, faster, stronger}, in:
  \bibinfo{booktitle}{Proceedings of the IEEE Conference on Computer Vision and
  Pattern Recognition (CVPR)}.
\bibitem[{Redmon and Farhadi(2018)}]{YOLOv3}
\bibinfo{author}{Redmon, J.}, \bibinfo{author}{Farhadi, A.},
  \bibinfo{year}{2018}.
\newblock \bibinfo{title}{Yolov3: An incremental improvement}.
\newblock \bibinfo{journal}{CoRR} \bibinfo{volume}{abs/1804.02767}.
\newblock \URLprefix \url{http://arxiv.org/abs/1804.02767},
  \href{http://arxiv.org/abs/1804.02767}{\tt arXiv:1804.02767}.
\bibitem[{Ren et~al.(2017)Ren, He, Girshick and Sun}]{Faster-RCNN}
\bibinfo{author}{Ren, S.}, \bibinfo{author}{He, K.}, \bibinfo{author}{Girshick,
  R.}, \bibinfo{author}{Sun, J.}, \bibinfo{year}{2017}.
\newblock \bibinfo{title}{Faster r-cnn: Towards real-time object detection with
  region proposal networks}.
\newblock \bibinfo{journal}{IEEE Transactions on Pattern Analysis and Machine
  Intelligence} \bibinfo{volume}{39}, \bibinfo{pages}{1137--1149}.
\newblock \DOIprefix\doi{10.1109/TPAMI.2016.2577031}.
\bibitem[{Rukhovich et~al.(2021)Rukhovich, Sofiiuk, Galeev, Barinova and
  Konushin}]{IterDet}
\bibinfo{author}{Rukhovich, D.}, \bibinfo{author}{Sofiiuk, K.},
  \bibinfo{author}{Galeev, D.}, \bibinfo{author}{Barinova, O.},
  \bibinfo{author}{Konushin, A.}, \bibinfo{year}{2021}.
\newblock \bibinfo{title}{Iterdet: Iterative scheme for object detection in
  crowded environments}, in: \bibinfo{editor}{Torsello, A.},
  \bibinfo{editor}{Rossi, L.}, \bibinfo{editor}{Pelillo, M.},
  \bibinfo{editor}{Biggio, B.}, \bibinfo{editor}{Robles-Kelly, A.} (Eds.),
  \bibinfo{booktitle}{Structural, Syntactic, and Statistical Pattern
  Recognition}, \bibinfo{publisher}{Springer International Publishing},
  \bibinfo{address}{Cham}. pp. \bibinfo{pages}{344--354}.
\bibitem[{Shao et~al.(2018)Shao, Zhao, Li, Xiao, Yu, Zhang and
  Sun}]{CrowdHuman}
\bibinfo{author}{Shao, S.}, \bibinfo{author}{Zhao, Z.}, \bibinfo{author}{Li,
  B.}, \bibinfo{author}{Xiao, T.}, \bibinfo{author}{Yu, G.},
  \bibinfo{author}{Zhang, X.}, \bibinfo{author}{Sun, J.}, \bibinfo{year}{2018}.
\newblock \bibinfo{title}{Crowdhuman: {A} benchmark for detecting human in a
  crowd}.
\newblock \bibinfo{journal}{CoRR} \bibinfo{volume}{abs/1805.00123}.
\newblock \URLprefix \url{http://arxiv.org/abs/1805.00123},
  \href{http://arxiv.org/abs/1805.00123}{\tt arXiv:1805.00123}.
\bibitem[{Sun et~al.(2021)Sun, Zhang, Jiang, Kong, Xu, Zhan, Tomizuka, Li,
  Yuan, Wang and Luo}]{Sparse-RCNN}
\bibinfo{author}{Sun, P.}, \bibinfo{author}{Zhang, R.}, \bibinfo{author}{Jiang,
  Y.}, \bibinfo{author}{Kong, T.}, \bibinfo{author}{Xu, C.},
  \bibinfo{author}{Zhan, W.}, \bibinfo{author}{Tomizuka, M.},
  \bibinfo{author}{Li, L.}, \bibinfo{author}{Yuan, Z.}, \bibinfo{author}{Wang,
  C.}, \bibinfo{author}{Luo, P.}, \bibinfo{year}{2021}.
\newblock \bibinfo{title}{Sparse r-cnn: End-to-end object detection with
  learnable proposals}, in: \bibinfo{booktitle}{2021 IEEE/CVF Conference on
  Computer Vision and Pattern Recognition (CVPR)}, pp.
  \bibinfo{pages}{14449--14458}.
\newblock \DOIprefix\doi{10.1109/CVPR46437.2021.01422}.
\bibitem[{Tian et~al.(2015)Tian, Luo, Wang and
  Tang}]{Deep-Learning-Strong-Parts-for-Pedestrian-Detection}
\bibinfo{author}{Tian, Y.}, \bibinfo{author}{Luo, P.}, \bibinfo{author}{Wang,
  X.}, \bibinfo{author}{Tang, X.}, \bibinfo{year}{2015}.
\newblock \bibinfo{title}{Deep learning strong parts for pedestrian detection},
  in: \bibinfo{booktitle}{2015 IEEE International Conference on Computer Vision
  (ICCV)}, pp. \bibinfo{pages}{1904--1912}.
\newblock \DOIprefix\doi{10.1109/ICCV.2015.221}.
\bibitem[{Tian et~al.(2019)Tian, Shen, Chen and He}]{FCOS}
\bibinfo{author}{Tian, Z.}, \bibinfo{author}{Shen, C.}, \bibinfo{author}{Chen,
  H.}, \bibinfo{author}{He, T.}, \bibinfo{year}{2019}.
\newblock \bibinfo{title}{Fcos: Fully convolutional one-stage object
  detection}, in: \bibinfo{booktitle}{2019 IEEE/CVF International Conference on
  Computer Vision (ICCV)}, pp. \bibinfo{pages}{9626--9635}.
\newblock \DOIprefix\doi{10.1109/ICCV.2019.00972}.
\bibitem[{Uijlings et~al.(2013)Uijlings, van~de Sande, Gevers and
  Smeulders}]{Selective-Search}
\bibinfo{author}{Uijlings, J.R.R.}, \bibinfo{author}{van~de Sande, K.E.A.},
  \bibinfo{author}{Gevers, T.}, \bibinfo{author}{Smeulders, A.W.M.},
  \bibinfo{year}{2013}.
\newblock \bibinfo{title}{Selective search for object recognition}.
\newblock \bibinfo{journal}{International Journal of Computer Vision}
  \bibinfo{volume}{104}, \bibinfo{pages}{154--171}.
\newblock \URLprefix
  \url{https://ivi.fnwi.uva.nl/isis/publications/2013/UijlingsIJCV2013}.
\bibitem[{Wang et~al.(2020)Wang, Bochkovskiy and Liao}]{Scaled-YOLOv4}
\bibinfo{author}{Wang, C.Y.}, \bibinfo{author}{Bochkovskiy, A.},
  \bibinfo{author}{Liao, H.Y.M.}, \bibinfo{year}{2020}.
\newblock \bibinfo{title}{Scaled-yolov4: Scaling cross stage partial network}.
\newblock \bibinfo{journal}{CoRR} \bibinfo{volume}{abs/2011.08036}.
\newblock \URLprefix \url{https://arxiv.org/abs/2011.08036},
  \href{http://arxiv.org/abs/2011.08036}{\tt arXiv:2011.08036}.
\bibitem[{Wang et~al.(2021)Wang, Song, Li, Sun, Sun and Zheng}]{POTO}
\bibinfo{author}{Wang, J.}, \bibinfo{author}{Song, L.}, \bibinfo{author}{Li,
  Z.}, \bibinfo{author}{Sun, H.}, \bibinfo{author}{Sun, J.},
  \bibinfo{author}{Zheng, N.}, \bibinfo{year}{2021}.
\newblock \bibinfo{title}{End-to-end object detection with fully convolutional
  network}, in: \bibinfo{booktitle}{2021 IEEE/CVF Conference on Computer Vision
  and Pattern Recognition (CVPR)}, pp. \bibinfo{pages}{15844--15853}.
\newblock \DOIprefix\doi{10.1109/CVPR46437.2021.01559}.
\bibitem[{Wang et~al.(2018)Wang, Xiao, Jiang, Shao, Sun and
  Shen}]{Repulsion-Loss}
\bibinfo{author}{Wang, X.}, \bibinfo{author}{Xiao, T.}, \bibinfo{author}{Jiang,
  Y.}, \bibinfo{author}{Shao, S.}, \bibinfo{author}{Sun, J.},
  \bibinfo{author}{Shen, C.}, \bibinfo{year}{2018}.
\newblock \bibinfo{title}{Repulsion loss: Detecting pedestrians in a crowd},
  in: \bibinfo{booktitle}{2018 IEEE/CVF Conference on Computer Vision and
  Pattern Recognition}, pp. \bibinfo{pages}{7774--7783}.
\newblock \DOIprefix\doi{10.1109/CVPR.2018.00811}.
\bibitem[{Xu et~al.(2020)Xu, Li, Yuan and Dang}]{Beta-RCNN}
\bibinfo{author}{Xu, Z.}, \bibinfo{author}{Li, B.}, \bibinfo{author}{Yuan, Y.},
  \bibinfo{author}{Dang, A.}, \bibinfo{year}{2020}.
\newblock \bibinfo{title}{Beta r-cnn: Looking into pedestrian detection from
  another perspective}, in: \bibinfo{booktitle}{Proceedings of the 34th
  International Conference on Neural Information Processing Systems},
  \bibinfo{publisher}{Curran Associates Inc.}, \bibinfo{address}{Red Hook, NY,
  USA}.
\bibitem[{Zhang et~al.(2017)Zhang, Benenson and Schiele}]{CityPersons}
\bibinfo{author}{Zhang, S.}, \bibinfo{author}{Benenson, R.},
  \bibinfo{author}{Schiele, B.}, \bibinfo{year}{2017}.
\newblock \bibinfo{title}{Citypersons: A diverse dataset for pedestrian
  detection}, in: \bibinfo{booktitle}{2017 IEEE Conference on Computer Vision
  and Pattern Recognition (CVPR)}, pp. \bibinfo{pages}{4457--4465}.
\newblock \DOIprefix\doi{10.1109/CVPR.2017.474}.
\bibitem[{Zhang et~al.(2021)Zhang, Chen, Yang and Schiele}]{Guided-Attention}
\bibinfo{author}{Zhang, S.}, \bibinfo{author}{Chen, D.}, \bibinfo{author}{Yang,
  J.}, \bibinfo{author}{Schiele, B.}, \bibinfo{year}{2021}.
\newblock \bibinfo{title}{Guided attention in cnns for occluded pedestrian
  detection and re-identification}.
\newblock \bibinfo{journal}{Int. J. Comput. Vision} \bibinfo{volume}{129},
  \bibinfo{pages}{1875每1892}.
\newblock \URLprefix \url{https://doi.org/10.1007/s11263-021-01461-z},
  \DOIprefix\doi{10.1007/s11263-021-01461-z}.
\bibitem[{Zhang et~al.(2020)Zhang, Chi, Yao, Lei and Li}]{ATSS}
\bibinfo{author}{Zhang, S.}, \bibinfo{author}{Chi, C.}, \bibinfo{author}{Yao,
  Y.}, \bibinfo{author}{Lei, Z.}, \bibinfo{author}{Li, S.Z.},
  \bibinfo{year}{2020}.
\newblock \bibinfo{title}{Bridging the gap between anchor-based and anchor-free
  detection via adaptive training sample selection}, in:
  \bibinfo{booktitle}{2020 IEEE/CVF Conference on Computer Vision and Pattern
  Recognition (CVPR)}, pp. \bibinfo{pages}{9756--9765}.
\newblock \DOIprefix\doi{10.1109/CVPR42600.2020.00978}.
\bibitem[{Zhang et~al.(2018)Zhang, Wen, Bian, Lei and Li}]{Occlusion-Aware}
\bibinfo{author}{Zhang, S.}, \bibinfo{author}{Wen, L.}, \bibinfo{author}{Bian,
  X.}, \bibinfo{author}{Lei, Z.}, \bibinfo{author}{Li, S.Z.},
  \bibinfo{year}{2018}.
\newblock \bibinfo{title}{Occlusion-aware r-cnn: Detecting pedestrians in a
  crowd}, in: \bibinfo{editor}{Ferrari, V.}, \bibinfo{editor}{Hebert, M.},
  \bibinfo{editor}{Sminchisescu, C.}, \bibinfo{editor}{Weiss, Y.} (Eds.),
  \bibinfo{booktitle}{Computer Vision -- ECCV 2018},
  \bibinfo{publisher}{Springer International Publishing},
  \bibinfo{address}{Cham}. pp. \bibinfo{pages}{657--674}.
\bibitem[{Zhao et~al.(2020)Zhao, Zhong, Fu, Tang and
  Pecht}]{Shrinkage-Networks}
\bibinfo{author}{Zhao, M.}, \bibinfo{author}{Zhong, S.}, \bibinfo{author}{Fu,
  X.}, \bibinfo{author}{Tang, B.}, \bibinfo{author}{Pecht, M.},
  \bibinfo{year}{2020}.
\newblock \bibinfo{title}{Deep residual shrinkage networks for fault
  diagnosis}.
\newblock \bibinfo{journal}{IEEE Transactions on Industrial Informatics}
  \bibinfo{volume}{16}, \bibinfo{pages}{4681--4690}.
\newblock \DOIprefix\doi{10.1109/TII.2019.2943898}.
\bibitem[{Zheng et~al.(2022)Zheng, Zhang, Zhang, Qi and Sun}]{Iter-E2EDET}
\bibinfo{author}{Zheng, A.}, \bibinfo{author}{Zhang, Y.},
  \bibinfo{author}{Zhang, X.}, \bibinfo{author}{Qi, X.}, \bibinfo{author}{Sun,
  J.}, \bibinfo{year}{2022}.
\newblock \bibinfo{title}{Progressive end-to-end object detection in crowded
  scenes}, in: \bibinfo{booktitle}{2022 IEEE/CVF Conference on Computer Vision
  and Pattern Recognition (CVPR)}, pp. \bibinfo{pages}{847--856}.
\newblock \DOIprefix\doi{10.1109/CVPR52688.2022.00093}.
\bibitem[{Zhou and
  Yuan(2017a)}]{Learning-to-Integrate-Occlusion-Specific-Detectors-for-Heavily-Occluded-Pedestrian-Detection}
\bibinfo{author}{Zhou, C.}, \bibinfo{author}{Yuan, J.}, \bibinfo{year}{2017}a.
\newblock \bibinfo{title}{Learning to integrate occlusion-specific detectors
  for heavily occluded pedestrian detection}, in: \bibinfo{editor}{Lai, S.H.},
  \bibinfo{editor}{Lepetit, V.}, \bibinfo{editor}{Nishino, K.},
  \bibinfo{editor}{Sato, Y.} (Eds.), \bibinfo{booktitle}{Computer Vision --
  ACCV 2016}, \bibinfo{publisher}{Springer International Publishing},
  \bibinfo{address}{Cham}. pp. \bibinfo{pages}{305--320}.
\bibitem[{Zhou and
  Yuan(2017b)}]{Multi-label-Learning-of-Part-Detectors-for-Heavily-Occluded-Pedestrian-Detection}
\bibinfo{author}{Zhou, C.}, \bibinfo{author}{Yuan, J.}, \bibinfo{year}{2017}b.
\newblock \bibinfo{title}{Multi-label learning of part detectors for heavily
  occluded pedestrian detection}, in: \bibinfo{booktitle}{2017 IEEE
  International Conference on Computer Vision (ICCV)}, pp.
  \bibinfo{pages}{3506--3515}.
\newblock \DOIprefix\doi{10.1109/ICCV.2017.377}.
\bibitem[{Zhou and Yuan(2018)}]{Bi-box}
\bibinfo{author}{Zhou, C.}, \bibinfo{author}{Yuan, J.}, \bibinfo{year}{2018}.
\newblock \bibinfo{title}{Bi-box regression for pedestrian detection and
  occlusion estimation}, in: \bibinfo{editor}{Ferrari, V.},
  \bibinfo{editor}{Hebert, M.}, \bibinfo{editor}{Sminchisescu, C.},
  \bibinfo{editor}{Weiss, Y.} (Eds.), \bibinfo{booktitle}{Computer Vision --
  ECCV 2018}, \bibinfo{publisher}{Springer International Publishing},
  \bibinfo{address}{Cham}. pp. \bibinfo{pages}{138--154}.
\bibitem[{Zhou et~al.(2020)Zhou, Zhou, Peng, Du, Sun, Guo and Huang}]{NOH-NMS}
\bibinfo{author}{Zhou, P.}, \bibinfo{author}{Zhou, C.}, \bibinfo{author}{Peng,
  P.}, \bibinfo{author}{Du, J.}, \bibinfo{author}{Sun, X.},
  \bibinfo{author}{Guo, X.}, \bibinfo{author}{Huang, F.}, \bibinfo{year}{2020}.
\newblock \bibinfo{title}{Noh-nms: Improving pedestrian detection by nearby
  objects hallucination}, in: \bibinfo{booktitle}{Proceedings of the 28th ACM
  International Conference on Multimedia}, \bibinfo{publisher}{Association for
  Computing Machinery}, \bibinfo{address}{New York, NY, USA}. p.
  \bibinfo{pages}{1967每1975}.
\newblock \URLprefix \url{https://doi.org/10.1145/3394171.3413617},
  \DOIprefix\doi{10.1145/3394171.3413617}.
\bibitem[{Zhou et~al.(2019)Zhou, Wang and Kr{\"{a}}henb{\"{u}}hl}]{CenterNet}
\bibinfo{author}{Zhou, X.}, \bibinfo{author}{Wang, D.},
  \bibinfo{author}{Kr{\"{a}}henb{\"{u}}hl, P.}, \bibinfo{year}{2019}.
\newblock \bibinfo{title}{Objects as points}.
\newblock \bibinfo{journal}{CoRR} \bibinfo{volume}{abs/1904.07850}.
\newblock \URLprefix \url{http://arxiv.org/abs/1904.07850},
  \href{http://arxiv.org/abs/1904.07850}{\tt arXiv:1904.07850}.
\bibitem[{Zhu et~al.(2020)Zhu, Su, Lu, Li, Wang and Dai}]{D-DETR}
\bibinfo{author}{Zhu, X.}, \bibinfo{author}{Su, W.}, \bibinfo{author}{Lu, L.},
  \bibinfo{author}{Li, B.}, \bibinfo{author}{Wang, X.}, \bibinfo{author}{Dai,
  J.}, \bibinfo{year}{2020}.
\newblock \bibinfo{title}{Deformable {DETR:} deformable transformers for
  end-to-end object detection}.
\newblock \bibinfo{journal}{CoRR} \bibinfo{volume}{abs/2010.04159}.
\newblock \URLprefix \url{https://arxiv.org/abs/2010.04159},
  \href{http://arxiv.org/abs/2010.04159}{\tt arXiv:2010.04159}.
\bibitem[{Zou et~al.(2020)Zou, Yang, Zhang and
  Ye}]{Attention-guided-neural-network}
\bibinfo{author}{Zou, T.}, \bibinfo{author}{Yang, S.}, \bibinfo{author}{Zhang,
  Y.}, \bibinfo{author}{Ye, M.}, \bibinfo{year}{2020}.
\newblock \bibinfo{title}{Attention guided neural network models for occluded
  pedestrian detection}.
\newblock \bibinfo{journal}{Pattern Recognition Letters} \bibinfo{volume}{131},
  \bibinfo{pages}{91--97}.
\newblock \URLprefix
  \url{https://www.sciencedirect.com/science/article/pii/S0167865519303733},
  \DOIprefix\doi{https://doi.org/10.1016/j.patrec.2019.12.010}.

\end{thebibliography}

\end{document}